\documentclass[runningheads]{llncs}
\usepackage{amsfonts}
\usepackage{graphicx}
\usepackage{amsfonts}
\usepackage{graphicx}
\usepackage{epsfig}
\usepackage{graphicx}
\usepackage{amsmath}
\usepackage{amssymb}
\usepackage{mathtools}
\usepackage{resizegather}
\usepackage{marvosym}
\usepackage{dsfont}
\usepackage{enumitem}
\usepackage{caption}
\usepackage{multirow}
\RequirePackage[utf8]{inputenc}
\usepackage{multirow}

\usepackage[colorlinks=true, allcolors=blue]{hyperref}

\usepackage[ruled, vlined]{algorithm2e}

\usepackage{lipsum}
\usepackage{array}
\usepackage{amsmath}

\usepackage{enumitem}
 
\usepackage{footnote}
\usepackage[hang, flushmargin]{footmisc}
\usepackage{footnotebackref}
\begin{document}

\title{Incremental Learning for Heterogeneous Structure Segmentation in Brain Tumor MRI}
\titlerunning{Incremental Learning for Heterogeneous Structure Segmentation}

\author{Xiaofeng Liu\inst{1} \and Helen A. Shih\inst{2}, Fangxu Xing\inst{1} \and  Emiliano Santarnecchi\inst{1} \and Georges El Fakhri\inst{1}  \and Jonghye Woo\inst{1}}

\institute{Gordon Center for Medical Imaging, Department of Radiology, Massachusetts General Hospital and Harvard Medical School, Boston, MA, 02114\and
Department of Radiation Oncology, Massachusetts General Hospital and Harvard Medical School, Boston, MA, 02114}

\authorrunning{X. Liu et al.}

\maketitle              

 
\begin{abstract}
Deep learning (DL) models for segmenting various anatomical structures have achieved great success via a static DL model that is trained in a single source domain. Yet, the static DL model is likely to perform poorly in a continually evolving environment, requiring appropriate model updates. In an incremental learning setting, we would expect that well-trained static models are updated, following continually evolving target domain data---e.g., additional lesions or structures of interest---collected from different sites, without catastrophic forgetting. This, however, poses challenges, due to distribution shifts, additional structures not seen during the initial model training, and the absence of training data in a source domain. To address these challenges, in this work, we seek to progressively evolve an ``off-the-shelf" trained segmentation model to diverse datasets with additional anatomical categories in a unified manner. Specifically, we first propose a divergence-aware dual-flow module with balanced rigidity and plasticity branches to decouple old and new tasks, which is guided by continuous batch renormalization. Then, a complementary pseudo-label training scheme with self-entropy regularized momentum MixUp decay is developed for adaptive network optimization. We evaluated our framework on a brain tumor segmentation task with continually changing target domains---i.e., new MRI scanners/modalities with incremental structures. Our framework was able to well retain the discriminability of previously learned structures, hence enabling the realistic life-long segmentation model extension along with the widespread accumulation of big medical data.



\end{abstract}

\section{Introduction} 
 
Accurate segmentation of a variety of anatomical structures is a crucial prerequisite for subsequent diagnosis or treatment~\cite{shusharina2020automated}. While recent advances in data-driven deep learning (DL) have achieved superior segmentation performance~\cite{tajbakhsh2020embracing}, the segmentation task is often constrained by the availability of costly pixel-wise labeled training datasets. In addition, even if static DL models are trained with extraordinarily large amounts of training datasets in a supervised learning manner~\cite{tajbakhsh2020embracing}, there exists a need for a segmentor to update a trained model with new data alongside incremental anatomical structures~\cite{liu2022deep}. 

In real-world scenarios, clinical databases are often sequentially constructed from various clinical sites with varying imaging protocols \cite{liu2023attentive,liu2023memory,liu2022act,liu2022subtype}. As well, labeled anatomical structures are incrementally increased with additional lesions or new structures of interest, depending on study goals or clinical needs~\cite{ozdemir2018learn,liu2022learning}. Furthermore, access to previously used data for training can be restricted, due to data privacy protocols \cite{liu2022learning,li2022domain}. Therefore, efficiently utilizing heterogeneous structure-incremental (HSI) learning is highly desired for clinical practice to develop a DL model that can be generalized well for different types of input data and varying structures involved. Straightforwardly fine-tuning DL models with either new structures~\cite{yang2022uncertainty} or heterogeneous data~\cite{li2022domain} in the absence of the data used for the initial model training, unfortunately, can easily overwrite previously learned knowledge, i.e., catastrophic forgetting \cite{yang2022uncertainty,li2022domain,kim2019incremental}.



At present, satisfactory methods applied in the realistic HSI setting are largely unavailable.
$First$, recent structure-incremental works cannot deal with domain shift. Early attempts~\cite{ozdemir2018learn} simply used exemplar data in the previous stage.~\cite{douillard2021plop,yu2022self,yang2022uncertainty,liu2022learning} combined a trained model prediction and a new class mask as a pseudo-label. However, predictions from the old model under a domain shift are likely to be unreliable \cite{zou2019confidence}. The widely used pooled feature statistics consistency~\cite{douillard2021plop,yang2022uncertainty} is also not applicable for heterogeneous data, since the statistics are domain-specific \cite{chang2019domain}. In addition, a few works~\cite{kanakis2020reparameterizing,liu2021adaptive,zhang2022representation} proposed to increase the capacity of networks to avoid directly overwriting parameters that are entangled with old and new knowledge. However, the solutions cannot be domain adaptive. $Second$, from the perspective of continuous domain adaptation with the consistent class label, old exemplars have been used for the application of prostate MRI segmentation \cite{you2022incremental}. While Li et al.~\cite{li2022domain} further proposed to recover the missing old stage data with an additional generative model, hallucinating realistic data, given only the trained model itself, is a highly challenging task~\cite{yin2020dreaming} and may lead to sensitive information leakage \cite{zhang2021unsupervised}. $Third$, while, for natural image classification, Kundu et al.~\cite{kundu2020class} updated the model for class-incremental unsupervised domain adaption, its class prototype is not applicable for segmentation. 


In this work, we propose a unified HSI segmentor evolving framework with a divergence-aware decoupled dual-flow (D$^3$F) module, which is adaptively optimized via HSI pseudo-label distillation using a momentum MixUp decay (MMD) scheme. To explicitly avoid the overwriting of previously learned parameters, our D$^3$F follows a ``divide-and-conquer" strategy to balance the old and new tasks with a fixed rigidity branch and a compensated learnable plasticity branch, which is guided by our novel divergence-aware continuous batch renormalization (cBRN). The complementary knowledge can be flexibly integrated with the model re-parameterization \cite{ding2021repvgg}. Our additional parameters are constant in training, and 0 in testing. Then, the flexible D$^3$F module is trained following the knowledge distillation with novel HSI pseudo-labels. Specifically, inspired by the self-knowledge distillation~\cite{kim2020self} and self-training~\cite{zou2019confidence} that utilize the previous prediction for better generalization, we adaptively construct the HSI pseudo-label with an MMD scheme to smoothly adjust the contribution of potential noisy old model predictions on heterogeneous data and progressively learned new model predictions along with the training. In addition, unsupervised self-entropy minimization is added to further enhance performance. 

Our main contributions can be summarized as follow:

$\bullet$ To our knowledge, this is the first attempt at realistic HSI segmentation with both incremental structures of interest and diverse domains.

$\bullet$ We propose a divergence-aware decoupled dual-flow module guided by our novel continuous batch renormalization (cBRN) for alleviating the catastrophic forgetting under domain shift scenarios.

  
$\bullet$ The adaptively constructed HSI pseudo-label with self-training is developed for efficient HSI knowledge distillation.

We evaluated our framework on anatomical structure segmentation tasks from different types of MRI data collected from multiple sites. Our HSI scheme demonstrated superior performance in segmenting all structures with diverse data distributions, surpassing conventional class-incremental methods without considering data shift, by a large margin.




 



\section{Methodology} 
 
For the segmentation model under incremental structures of interest and domain shift scenarios, we are given an off-the-shelf segmentor $f_{\theta^0}:\mathcal{X}^0\rightarrow \mathcal{Y}^0$ parameterized with $\theta^0$, which has been trained with the data $\{x_n^0,y_n^0\}_{n=1}^{N^0}$ in an initial source domain $\mathcal{D}^0=\{\mathcal{X}^0,\mathcal{Y}^0\}$, where $x_n^0\in\mathbb{R}^{H\times W}$ and $y_n^0\in\mathbb{R}^{H\times W}$ are the paired image slice and its segmentation mask with the height of $H$ and width of $W$, respectively. There are $T$ consecutive evolving stages with heterogeneous target domains $\mathcal{D}^t=\{\mathcal{X}^t,\mathcal{S}^t\}_{t=1}^T$, each with the paired slice set $\{x_n^t\}_{n=1}^{N^t}\in\mathcal{X}^t$ and the current stage label set $\{
s_n^t\}_{n=1}^{N^t}\in\mathcal{S}^t$, where $x_n^t, s_n^t\in\mathbb{R}^{H\times W}$. Due to heterogeneous domain shifts, $\mathcal{X}^t$ from different sites or modalities follows diverse distributions across all $T$ stages. Due to incremental anatomical structures, the overall label space, across the previous $t$ stages, $\mathcal{Y}^t$ is expanded from $\mathcal{Y}^{t-1}$ with the additional annotated structures $\mathcal{S}^t$ in stage $t$, i.e., $\mathcal{Y}^t=\mathcal{Y}^{t-1}\cup\mathcal{S}^t=\mathcal{Y}^0\cup\mathcal{S}^1\cdots\cup\mathcal{S}^t$. We are targeting to learn $f_{\theta^T}:\{\mathcal{X}^t\}_{t=1}^T\rightarrow \mathcal{Y}^T$ that performs well on all $\{\mathcal{X}^t\}_{t=1}^T$ for delineating all of the structures $\mathcal{Y}^T$ seen in $T$ stages.

\begin{figure*}[t]
\begin{center}\label{img1} 
\includegraphics[width=1\linewidth]{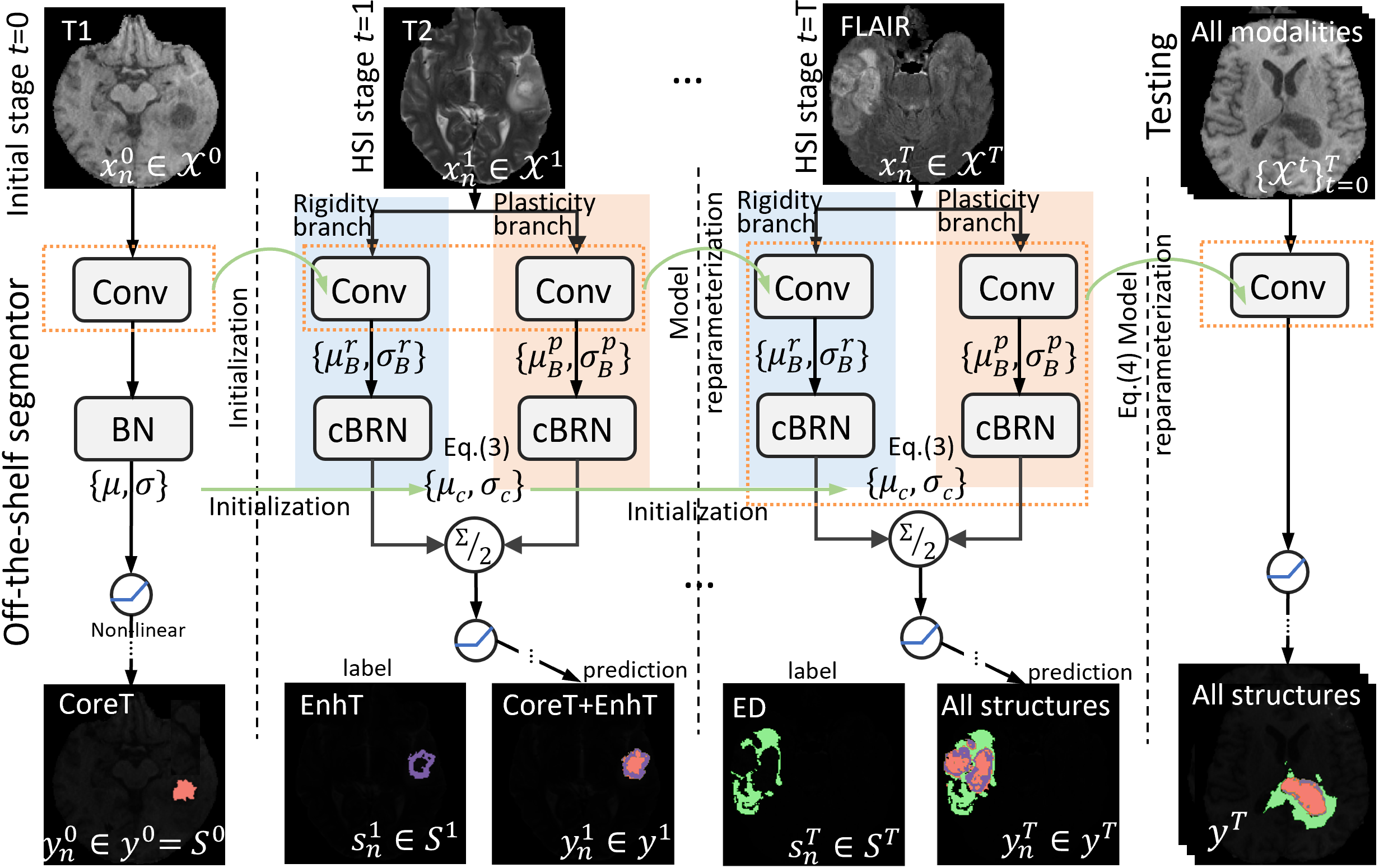}
\end{center}   
\caption{Illustration of one layer in our proposed divergence-aware decoupled dual-flow module guided with cBRN for our cross-MR-modality HSI task, i.e., subject-independent (CoreT with T1) $\rightarrow$ (EnhT with T2) $\rightarrow$ (ED with FLAIR). Notably, we do not require the dual-flow or cBRN, for the initial segmentor.}   
\label{method1}\end{figure*} 

\subsection{cBRN guided divergence-aware decoupled dual-flow}

To alleviate the forgetting through parameter overwriting, caused by both new structures and data shift, we propose a D$^3$F module for flexible decoupling and integration of old and new knowledge. 

Specifically, we duplicate the convolution in each layer initialized with the previous model $f_{\theta^{t-1}}$ to form two branches as in~\cite{kanakis2020reparameterizing,liu2021adaptive,zhang2022representation}. The first $rigidity$ branch $f_{\theta^t}^r$ is fixed at the stage $t$ to keep the old knowledge we have learned. In contrast, the extended $plasticity$ branch $f_{\theta^t}^p$ is expected to be adaptively updated to learn the new task in $\mathcal{D}^t$. At the end of current training stage $t$, we can flexibly integrate the convolutions in two branches, i.e., $\{W_t^r,b_t^r\}$ and $\{W_t^p,b_t^p\}$ to $\{W_{t+1}^r=\frac{W_t^r+W_t^p}{2},b_{t+1}^r=\frac{b_t^r+b_t^p}{2}\}$  with the model re-parameterization \cite{ding2021repvgg}. In fact, the dual-flow model can be regarded as an implicit ensemble scheme \cite{huang2016deep} to integrate multiple sub-modules with a different focus. In addition, as demonstrated in \cite{fu2021interactive}, the fixed modules will regularize the learnable modules to act as the fixed one. Thus, the plasticity modules can also be implicitly encouraged to keep the previous knowledge along with its HSI learning.

However, under the domain shift, it can be sub-optimal to directly average the parameters, since $f_{\theta^t}^r$ may not perform well to predict $\mathcal{Y}^{t-1}$ on $\mathcal{X}^t$. It has been demonstrated that batch statistics adaptation plays an important role in domain generalizable model training \cite{liu2021adapting}. Therefore, we propose a continual batch renormalization (cBRN) to mitigate the feature statistics divergence between each training batch at a specific stage and the life-long global data distribution. 


Of note, as a default block in the modern convolutional neural networks (CNN) \cite{He_2016_CVPR,zhou2019normalization}, batch normalization (BN) \cite{ioffe2015batch} normalizes the input feature of each CNN channel $z\in\mathbb{R}^{H_c\times W_c}$ with its batch-wise statistics, e.g., mean $\mu_B$ and standard deviation $\sigma_B$, and learnable scaling and shifting factors $\{\gamma,\beta\}$ as $\tilde{z}_i= \frac{z_i-\mu_B}{\sigma_B} \cdot \gamma+\beta,$ where $i$ indexes the spatial position in $\mathbb{R}^{H_c\times W_c}$. BN assumes that the same mini-batch training and testing distribution~\cite{ioffe2017batch}, which does not hold in HSI. Simply enforcing the same statistics across domains as~\cite{douillard2021plop,yu2022self,yang2022uncertainty} can weaken the model expressiveness \cite{zhang2020generalizable}.


The recent BRN~\cite{ioffe2017batch} proposes to rectify the data shift between each batch and the dataset by using the moving average ${\mu}$ and $\sigma$ along with the training: 
\begin{align}
\mu=(1-\eta)\cdot\mu+\eta\cdot\mu_B, ~~\sigma=(1-\eta)\cdot\sigma+\eta\cdot\sigma_B,
\end{align}

\noindent where $\eta\in[0,1]$ is applied to balance the global statistics and the current batch. In addition,
$\gamma=\frac{\sigma_B}{\sigma}$ and $\beta=\frac{\mu_B-\mu}{\sigma}$ are used in both training and testing. Therefore, BRN renormalizes $\tilde{z}_i= \frac{z_i-\mu}{\sigma}$ to highlight the dependency on the global statistics $\{\mu,\sigma\}$ in training for a more generalizable model, while limited to the static learning.


In this work, we further explore the potential of BRN in the continuously evolving HSI task to be general for all of domains involved. Specifically, we extend BRN to cBRN across multiple consecutive stages by updating $\{\mu_c,\sigma_c\}$ along with all stages of training, which is transferred as shown in Fig.~1. The conventional BN also inherits $\{\mu,\sigma\}$ for testing, while not being used in training~\cite{ioffe2015batch}. At the stage $t$, $\mu_c$ and $\sigma_c$ are succeeded from $t-1$ stage, and are updated with the current batch-wise $\{\mu_B^r,\sigma_B^r\}$ and $\{\mu_B^p,\sigma_B^p\}$ in rigidity and plasticity branches: 
\begin{align}
\mu_c=(1-\eta)\cdot\mu_c+\eta\cdot\frac{1}{2}\{\mu_B^r+\mu_B^p\}, ~~\sigma_c=(1-\eta)\cdot\sigma_c+\eta\cdot\frac{1}{2}\{\sigma_B^r+\sigma_B^p\}.
\end{align} 

For testing, the two branches in final model $f_{\theta^T}$ can be merged for the lightweight implementation: 
\begin{align}
\tilde{z}=\frac{W_T^r z+b_T^r+\mu_c}{2\sigma_c}+\frac{W_T^p z+b_T^p+\mu_c}{2\sigma_c}=\frac{W_T^r+W_T^p}{2\sigma_c}z+\frac{b_T^r+b_T^p-2\mu_c}{2\sigma_c}=\hat{W}z+\hat{b}.
\end{align}

\noindent Therefore, $f_{\theta}^T$ does not introduce additional parameters for deployment. 

\begin{figure*}[t]
\begin{center}\label{img2}
\includegraphics[width=1\linewidth]{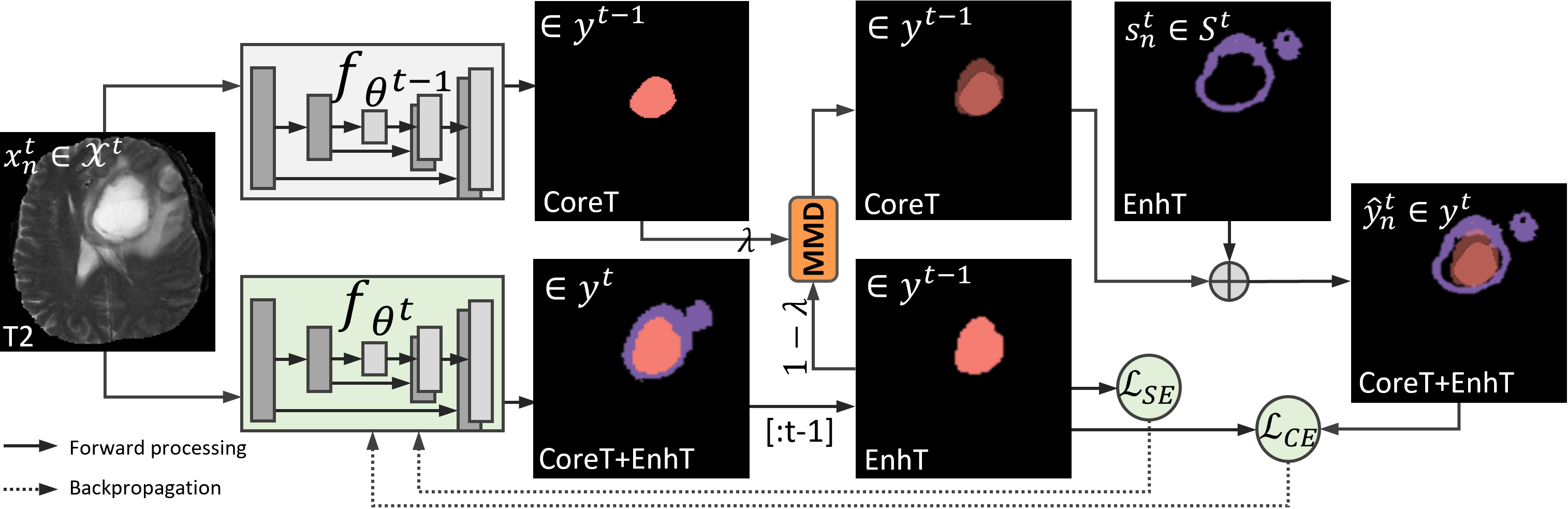}
\end{center}  
\caption{Illustration of the proposed HSI pseudo-label distillation with MMD}    
\label{method2}\end{figure*}

\subsection{HSI pseudo-label distillation with momentum MixUp decay} 

The training of our developed $f_{\theta^t}$ with D$^3$F is supervised with the previous model $f_{\theta^{t-1}}$ and current stage data $\{x_n^t,s_n^t\}_{n=1}^{N^t}$. In conventional class incremental learning, the knowledge distillation \cite{yin2020dreaming} is widely used to construct the combined label ${y}_n^t\in\mathbb{R}^{H\times W}$ by adding $s_n^t$ and the prediction of $f_{\theta^{t-1}}(x_n^t)$. Then, $f_{\theta^t}$ can be optimized by the training pairs of $\{x_n^t,{y}_n^t\}_{n=1}^{N^t}$. However, with heterogeneous data in different stages, $f_{\theta^{t-1}}(x_n^t)$ can be highly unreliable. Simply using it as ground truth cannot guide the correct knowledge transfer.

In this work, we construct a complementary pseudo-label $\hat{y}_n^t\in\mathbb{R}^{H\times W}$ with a MixUp decay scheme to adaptively exploit the knowledge in the old segmentor for the progressively learned new segmentor. In the initial training epochs, $f_{\theta^{t-1}}$ could be a more reliable supervision signal, while we would expect $f_{\theta^{t}}$ can learn to perform better on predicting $\mathcal{Y}^{t-1}$. Of note, even with the rigidity branch, the integrated network can be largely distracted by the plasticity branch in the initial epochs. Therefore, we propose to dynamically adjust their importance in constructing pseudo-label along with the training progress. Specifically, we MixUp the predictions of $f_{\theta^{t-1}}$ and $f_{\theta^{t}}$ w.r.t. $\mathcal{Y}^{t-1}$, i.e., $f_{\theta^{t}}(\cdot)[:t-1]$, and control their pixel-wise proportion for the pseudo-label $\hat{y}_n^t$ with MMD:  
\begin{align}
\hat{y}_{n:i}^t = \{\lambda f_{\theta^{t-1}}(x_{n:i}^t) + (1-\lambda) f_{\theta^{t}}(x_{n:i}^t)[:t-1]\}\cup s_{n:i}^t, ~\lambda= \lambda^0\text{exp}(-I),
\end{align} 
where $i$ indexes each pixel, and $\lambda$ is the adaptation momentum factor with the exponential decay of iteration $I$. $\lambda^0$ is the initial weight of $f_{\theta^{t-1}}(x_{n:i}^t)$, which is empirically set to 1 to constrain $\lambda\in(0,1]$. Therefore, the weight of old model prediction can be smoothly decreased along with the training, and $f_{\theta^{t}}(x_{n:i}^t)$ gradually represents the target data for the old classes in $[:t-1]$. Of note, we have ground-truth of new structure $s_{n:i}^t$ under HSI scenarios \cite{douillard2021plop,yu2022self,yang2022uncertainty,liu2022learning}. We calculate the cross-entropy loss $\mathcal{L}_{CE}$ with the pseudo-label $\hat{y}_{n:i}^t$ as self-training~\cite{kim2020self,zou2019confidence}.

In addition to the old knowledge inherited in $f_{\theta^{t-1}}$, we propose to explore unsupervised learning protocols to stabilize the initial training. We adopt the widely used self-entropy (SE) minimization \cite{grandvalet2005semi} as a simple add-on training objective. Specifically, we have the slice-level segmentation SE, which is the averaged entropy of the pixel-wise softmax prediction as $ \mathcal{L}_{SE}=
    \mathbb{E}_i\{-{f_{\theta^{t}}(x_{n:i}^t) \text{log} f_{\theta^{t}}(x_{n:i}^t)}\}.$ In training, the overall optimization loss is formulated as follows:  
\begin{align}
    \mathcal{L}=\mathcal{L}_{CE}(\hat{y}_{n:i}^t,f_{\theta^{t}}(x_{n:i}^t))+\alpha\mathcal{L}_{SE}(f_{\theta^{t}}(x_{n:i}^t)),~~\alpha=\frac{I_{max}-I}{I_{max}}{\alpha^0},
\end{align}

\noindent where $\alpha$ is used to balance our HSI distillation and SE minimization terms, and $I_{max}$ is the scheduled iteration. Of note, strictly minimizing the SE can result in a trivial solution of always predicting a one-hot distribution \cite{grandvalet2005semi}, and a linear decreasing of $\alpha$ is usually applied, where $\lambda^0$ and $\alpha^0$ are reset in each stage.

\begin{table}[t] 
\caption{Numerical comparisons and ablation studies of the cross-subset brain tumor HSI segmentation task} 
\resizebox{1\linewidth}{!}{
\begin{tabular}{l|c|cccc|cccc}
\hline
\multirow{2}{*}{Method}                          & Data shift           & \multicolumn{4}{c|}{Dice similarity coefficient (DSC)  {[}\%{]} $\uparrow$}                                                                                             & \multicolumn{4}{c}{Hausdorff distance (HD){[}mm{]} $\downarrow$}                                                                                                        \\ \cline{3-10} 
                                                 & consideration           & \multicolumn{1}{c|}{~Mean~}                          & \multicolumn{1}{c|}{~CoreT~} & \multicolumn{1}{c|}{~EnhT~} & ~ED~ & \multicolumn{1}{c|}{~Mean~}                        & \multicolumn{1}{c|}{~CoreT~} & \multicolumn{1}{c|}{~EnhT~} & ~~ED~~ \\ \hline \hline
PLOP~\cite{douillard2021plop}    & $\times$             & \multicolumn{1}{c|}{59.83$\pm0.131$}                           & \multicolumn{1}{c|}{45.50}             & \multicolumn{1}{c|}{57.39}            & 76.59          & \multicolumn{1}{c|}{19.2$\pm$0.14}                           & \multicolumn{1}{c|}{22.0}              & \multicolumn{1}{c|}{19.8}             & 15.9                       \\ \hline
MargExcIL~\cite{liu2022learning} & $\times$             & \multicolumn{1}{c|}{60.49$\pm$0.127}                           & \multicolumn{1}{c|}{48.37}             & \multicolumn{1}{c|}{56.28}            & 76.81          & \multicolumn{1}{c|}{18.9$\pm$0.11}                           & \multicolumn{1}{c|}{21.4}              & \multicolumn{1}{c|}{19.8}             & 15.5                       \\ \hline
UCD~\cite{yang2022uncertainty}   & $\times$             & \multicolumn{1}{c|}{61.84$\pm$0.129}                           & \multicolumn{1}{c|}{49.23}             & \multicolumn{1}{c|}{58.81}            & 77.48          & \multicolumn{1}{c|}{19.0$\pm$0.15}                           & \multicolumn{1}{c|}{21.8}              & \multicolumn{1}{c|}{19.4}             & 15.7                       \\ \hline\hline
HSI-MMD                                          & $\surd$              & \multicolumn{1}{c|}{66.87$\pm$0.126}                           & \multicolumn{1}{c|}{59.42}             & \multicolumn{1}{c|}{61.26}            & 79.93          & \multicolumn{1}{c|}{16.8$\pm$0.13}                           & \multicolumn{1}{c|}{18.5}              & \multicolumn{1}{c|}{17.8}             & 14.2                       \\ \hline
HSI-D$^3$F                                       & $\surd$              & \multicolumn{1}{c|}{67.18$\pm$0.118}                           & \multicolumn{1}{c|}{60.18}             & \multicolumn{1}{c|}{63.09}            & 78.26          & \multicolumn{1}{c|}{16.7$\pm$0.14}                           & \multicolumn{1}{c|}{18.0}              & \multicolumn{1}{c|}{17.5}             & 14.5                       \\ \hline
HSI-cBRN                                         & $\surd$              & \multicolumn{1}{c|}{68.07$\pm$0.121}                           & \multicolumn{1}{c|}{61.52}             & \multicolumn{1}{c|}{63.45}            & 79.25          & \multicolumn{1}{c|}{16.3$\pm$0.14}                           & \multicolumn{1}{c|}{17.8}              & \multicolumn{1}{c|}{17.3}             & 13.8                       \\ \hline
\textbf{HSI}                    & $\surd$              & \multicolumn{1}{c|}{\textbf{69.44$\pm$0.119}} & \multicolumn{1}{c|}{\textbf{63.79}}             & \multicolumn{1}{c|}{\textbf{64.71}}            & \textbf{79.81}          & \multicolumn{1}{c|}{\textbf{15.7$\pm$0.12}} & \multicolumn{1}{c|}{\textbf{16.7}}              & \multicolumn{1}{c|}{\textbf{16.9}}             & \textbf{13.6}                       \\ \hline\hline
Joint Static                                     & $\surd$(upper bound) & \multicolumn{1}{c|}{73.98$\pm$0.117}                           & \multicolumn{1}{c|}{71.14}             & \multicolumn{1}{c|}{68.35}            & 82.46          & \multicolumn{1}{c|}{15.0$\pm$0.13}                           & \multicolumn{1}{c|}{15.7}              & \multicolumn{1}{c|}{16.2}             & 13.2                       \\ \hline
\end{tabular}\label{tab1}} 
\end{table}

\section{Experiments and Results}


We carried out two evaluation settings using the BraTS2018 database \cite{bakas2018identifying}, including cross-subset (relatively small domain shift) and cross-modality (relatively large domain shift) tasks. The BraTS2018 database is a continually evolving database \cite{bakas2018identifying} with a total of 285 glioblastoma or low-grade gliomas subjects, comprising three consecutive subsets, i.e., 30 subjects from BraTS2013 \cite{menze2014multimodal}, 167 subjects from TCIA \cite{clark2013cancer}, and 88 subjects from CBICA \cite{bakas2018identifying}. Notably, these three subsets were collected from different clinical sites, vendors, or populations \cite{bakas2018identifying}. Each subject has T1, T1ce, T2, and FLAIR MRI volumes with voxel-wise labels for the tumor core (CoreT), the enhancing tumor (EnhT), and the edema (ED).


We incrementally learned CoreT, EnhT, and ED structures throughout three consecutive stages, each following different data distributions. We used subject-independent 7/1/2 split for training, validation, and testing. For a fair comparison, we adopted the ResNet-based 2D nnU-Net backbone with BN as in \cite{isensee2021nnu} for all of the methods and all stages used in this work. 



\begin{figure*}[t]
\begin{center}  
\includegraphics[width=1\linewidth]{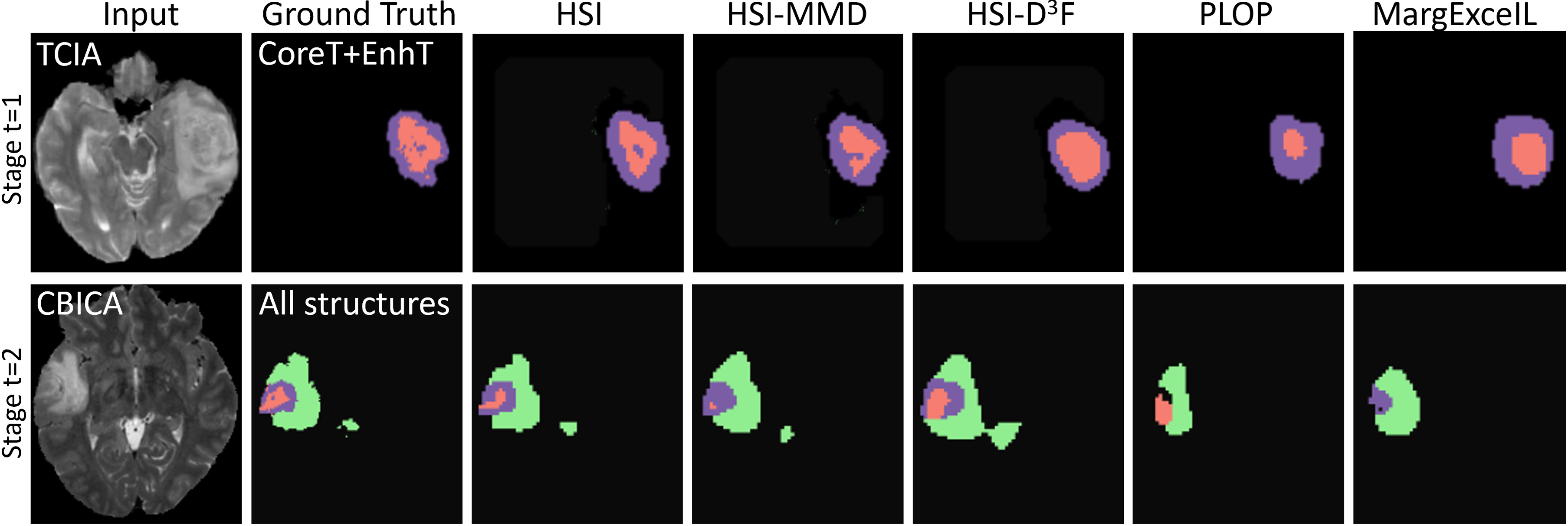}
\end{center}  
\caption{Segmentation examples in $t=1$ and $t=2$ in the cross-subset brain tumor HSI segmentation task.}  
\label{img3}\end{figure*}

\subsection{Cross-subset structure incremental evolving}
 
In our cross-subset setting, three structures were sequentially learned across three stages: (CoreT with BraTS2013) $\rightarrow$ (EnhT with TCIA) $\rightarrow$ (ED with CBICA). Of note, we used a CoreT segmentator trained with BraTS2013 as our off-the-shelf segmentor in $t=0$. Testing involved all subsets and anatomical structures. We compared our framework with the three typical structure-incremental (SI-only) segmentation methods, e.g., PLOP \cite{douillard2021plop}, MargExcIL \cite{liu2022learning}, and UCD \cite{yang2022uncertainty}, which cannot address the heterogeneous data across stages. As tabulated in Table~\ref{tab1}, PLOP \cite{douillard2021plop} with additional feature statistic constraints has lower performance than MargExcIL \cite{liu2022learning}, since the feature statistic consistency was not held in HSI scenarios. Of note, the domain-incremental methods \cite{li2022domain,you2022incremental} cannot handle the changing output space. Our proposed HSI framework outperformed SI-only methods \cite{douillard2021plop,liu2022learning,yang2022uncertainty} with respect to both DSC and HD, by a large margin. For the anatomical structure CoreT learned in $t=0$, the difference between our HSI and these SI-only methods was larger than 10\% DSC, which indicates the data shift related forgetting lead to a more severe performance drop in the early stages. We set $\eta=0.01$ and $\ alpha^0=10$ according to the sensitivity study in the supplementary material.


For the ablation study, we denote HSI-D$^3$F as our HSI without the D$^3$F module, simply fine-tuning the model parameters. HSI-cBRN used dual-flow to avoid direct overwriting, while the model was not guided by cBRN for more generalized prediction on heterogeneous data. As shown in Table \ref{tab1}, both the dual-flow and cBRN improve the performance. Notably, the dual-flow model with flexible re-parameterization was able to alleviate the overwriting, while our cBRN was developed to deal with heterogeneous data. In addition, HSI-MMD indicates our HSI without the momentum MixUp decay in pseudo-label construction, i.e., simply regarding the prediction of $f_{\theta^{t-1}}(x^t)$ is ground truth for $\mathcal{Y}^{t-1}$. However, $f_{\theta^{t-1}}(x^t)$ can be quite noisy, due to the low quantification performance of early stage structures, which can be aggravated in the case of the long-term evolving scenario. Of note, the pseudo-label construction is necessary as in~\cite{douillard2021plop,liu2022learning,yang2022uncertainty}. We also provide the qualitative comparison with SI-only methods and ablation studies in Fig. \ref{img3}.





\begin{table}[t]
\centering  
\caption{Numerical comparisons and ablation studies of the cross-modality brain tumor HSI segmentation task}
\resizebox{1\linewidth}{!}{
\begin{tabular}{l|c|cccc|cccc}
\hline
\multirow{2}{*}{Method}                          & Data shift            & \multicolumn{4}{c|}{Dice similarity coefficient (DSC)  {[}\%{]} $\uparrow$}                                                                                             & \multicolumn{4}{c}{Hausdorff distance (HD){[}mm{]} $\downarrow$}                                                                                                        \\ \cline{3-10} 
                                                 & consideration            & \multicolumn{1}{c|}{~Mean~}                          & \multicolumn{1}{c|}{~CoreT~} & \multicolumn{1}{c|}{~EnhT~} & ~ED~ & \multicolumn{1}{c|}{~Mean~}                        & \multicolumn{1}{c|}{~CoreT~} & \multicolumn{1}{c|}{~EnhT~} & ~~ED~~ \\ \hline \hline
PLOP~\cite{douillard2021plop}    & $\times$              & \multicolumn{1}{c|}{39.58$\pm$0.231}                           & \multicolumn{1}{c|}{13.84}             & \multicolumn{1}{c|}{38.93}            & 65.98          & \multicolumn{1}{c|}{30.7$\pm$0.26}                           & \multicolumn{1}{c|}{48.1}              & \multicolumn{1}{c|}{25.4}             & 18.7                       \\ \hline
MargExcIL~\cite{liu2022learning} & $\times$              & \multicolumn{1}{c|}{42.84$\pm$0.189}                           & \multicolumn{1}{c|}{19.56}             & \multicolumn{1}{c|}{41.56}            & 67.40          & \multicolumn{1}{c|}{29.1$\pm$0.28}                           & \multicolumn{1}{c|}{46.7}              & \multicolumn{1}{c|}{22.1}             & 18.6                       \\ \hline
UCD~\cite{yang2022uncertainty}   & $\times$              & \multicolumn{1}{c|}{44.67$\pm$0.214}                           & \multicolumn{1}{c|}{21.39}             & \multicolumn{1}{c|}{45.28}            & 67.35          & \multicolumn{1}{c|}{29.4$\pm$0.32}                           & \multicolumn{1}{c|}{46.2}              & \multicolumn{1}{c|}{23.6}             & 18.4                       \\ \hline \hline
HSI-MMD                                          & $\surd$               & \multicolumn{1}{c|}{59.81$\pm$0.207}                           & \multicolumn{1}{c|}{51.63}             & \multicolumn{1}{c|}{53.82}            & 73.97          & \multicolumn{1}{c|}{19.4$\pm$0.26}                           & \multicolumn{1}{c|}{21.6}              & \multicolumn{1}{c|}{20.5}             & 16.2                       \\ \hline
HSI-D$^3$F                                       & $\surd$               & \multicolumn{1}{c|}{60.81$\pm$0.195}                           & \multicolumn{1}{c|}{53.87}             & \multicolumn{1}{c|}{55.42}            & 73.15          & \multicolumn{1}{c|}{19.2$\pm$0.21}                           & \multicolumn{1}{c|}{21.4}              & \multicolumn{1}{c|}{19.9}             & 16.2                       \\ \hline
HSI-cBRN                                         & $\surd$               & \multicolumn{1}{c|}{61.87$\pm$0.180}                           & \multicolumn{1}{c|}{54.90}             & \multicolumn{1}{c|}{56.62}            & 74.08          & \multicolumn{1}{c|}{18.5$\pm$0.25}                           & \multicolumn{1}{c|}{20.1}              & \multicolumn{1}{c|}{19.5}             & 16.0                       \\ \hline
\textbf{HSI}                    & $\surd$               & \multicolumn{1}{c|}{\textbf{64.15$\pm$0.205}} & \multicolumn{1}{c|}{\textbf{58.11}}             & \multicolumn{1}{c|}{\textbf{59.51}}            & \textbf{74.83}          & \multicolumn{1}{c|}{\textbf{17.7$\pm$0.29}} & \multicolumn{1}{c|}{\textbf{18.9}}              & \multicolumn{1}{c|}{\textbf{18.6}}             & \textbf{15.8}                       \\ \hline \hline
Joint Static                                     & $\surd$ (upper bound) & \multicolumn{1}{c|}{70.64$\pm$0.184}                           & \multicolumn{1}{c|}{67.48}             & \multicolumn{1}{c|}{65.75}            & 78.68          & \multicolumn{1}{c|}{16.7$\pm$0.26}                           & \multicolumn{1}{c|}{17.2}              & \multicolumn{1}{c|}{17.8}             & 15.1                       \\ \hline
\end{tabular}\label{tab2}}  
\end{table}

\subsection{Cross-modality structure incremental evolving}
 
In our cross-modality setting, three structures were sequentially learned across three stages: (CoreT with T1) $\rightarrow$ (EnhT with T2) $\rightarrow$ (ED with T2 FLAIR). Of note, we used the CoreT segmentator trained with T1 modality as our off-the-shelf segmentor in $t=0$. Testing involved all MRI modalities and all structures. With the hyperparameter validation, we empirically set $\eta=0.01$ and $\alpha^0=10$. 

In Table \ref{tab2}, we provide quantitative evaluation results. We can see that our HSI framework outperformed SI-only methods \cite{douillard2021plop,liu2022learning,yang2022uncertainty} consistently. The improvement can be even larger, compared with the cross-subset task, since we have much more diverse input data in the cross-modality setting. Catastrophic forgetting can be severe, when we use SI-only method for predicting early stage structures, e.g., CoreT. We also provide the ablation study with respect to D$^3$F, cBRN, and MMD in Table \ref{tab2}. The inferior performance of HSI-D$^3$F/cBRN/MMD demonstrates the effectiveness of these modules for mitigating domain shifts. 


\section{Conclusion}
 
This work proposed an HSI framework under a clinically meaningful scenario, in which clinical databases are sequentially constructed from different sites/imaging protocols with new labels. To alleviate the catastrophic forgetting alongside continuously varying structures and data shifts, our HSI resorted to a D$^3$F module for learning and integrating old and new knowledge nimbly. In doing so, we were able to achieve divergence awareness with our cBRN-guided model adaptation for all the data involved. Our framework was optimized with a self-entropy regularized HSI pseudo-label distillation scheme with MMD to efficiently utilize the previous model in different types of MRI data. Our framework demonstrated superior segmentation performance in learning new anatomical structures from cross-subset/modality MRI data. It was experimentally shown that a large improvement in learning anatomic structures was observed. 


\section*{Acknowledgements}

This work is supported by NIH R01DC018511, R01DE027989, and P41EB022544. The authors would like to thank Dr. Jonghyun Choi for his valuable insights and helpful discussions.

\bibliographystyle{splncs04}
\bibliography{egbib}

\end{document}